\documentclass[10pt,twocolumn,letterpaper]{article}

\usepackage{amsmath}
\usepackage{graphicx}
\usepackage{amssymb}
\usepackage{bbm}
\usepackage{float}
\usepackage{bm}

\usepackage[pagenumbers]{cvpr} 

\usepackage[dvipsnames]{xcolor}

\definecolor{cvprblue}{rgb}{0.21,0.49,0.74}
\usepackage[pagebackref,breaklinks,colorlinks,citecolor=cvprblue]{hyperref}

\def\paperID{*****} 
\def\confName{CVPR}
\def\confYear{2024}

\title{DiVE: DiT-based Video Generation with Enhanced Control}

\author{
Junpeng Jiang$^{1,2}$\footnotemark[1], Gangyi Hong$^{3,2}$\footnotemark[1], Lijun Zhou$^{2}$, Enhui Ma$^{4,2}$, Hengtong Hu$^{2}$, Xia Zhou$^{2}$\\
Jie Xiang$^{2}$, Fan Liu$^{5}$, Kaicheng Yu$^{4}$, Haiyang Sun$^{2}$, Kun Zhan$^{2}$, Peng Jia$^{2}$, Miao Zhang$^{1}$\footnotemark[2]\\
$^{1}$Harbin Institute of Technology (Shenzhen) \quad  $^{2}$Li Auto Inc. \quad  $^{3}$Tsinghua University\\
$^{4}$Westlake University \quad $^{5}$National University of Singapore \\
\textit{Project Page}:~\href{https://liautoad.github.io/DIVE/}{https://liautoad.github.io/DIVE/}
}

\begin{document}
\maketitle

\renewcommand{\thefootnote}{\fnsymbol{footnote}}
\footnotetext[1]{Equal contribution.}
\footnotetext[2]{Corresponding author.}
\renewcommand{\thefootnote}{\arabic{footnote}}

\begin{abstract}
Generating high-fidelity, temporally consistent videos in autonomous driving scenarios faces a significant challenge, \textit{e.g.} problematic maneuvers in corner cases. Despite recent video generation works are proposed to tackcle the mentioned problem, \textit{i.e.} models built on top of Diffusion Transformers (DiT), works are still missing which are targeted on exploring the potential for multi-view videos generation scenarios. 
Noticeably, we propose the first DiT-based framework specifically designed for generating temporally and multi-view consistent videos which precisely match the given bird's-eye view layouts control. 
Specifically, the proposed framework leverages a parameter-free spatial view-inflated attention mechanism to guarantee the cross-view consistency, where joint cross-attention modules and ControlNet-Transformer are integrated to further improve the precision of control. 
To demonstrate our advantages, we extensively investigate the qualitative comparisons on nuScenes dataset, particularly in some most challenging corner cases. 
In summary, the effectiveness of our proposed method in producing long, controllable, and highly consistent videos under difficult conditions is proven to be effective.
\end{abstract}

\vspace{-12pt}    
\section{Introduction}
\label{sec:intro}

Bird's-Eye-View (BEV) perception has gained significant attention for autonomous driving, highlighting its immense potential in tasks such as 3D object detection \cite{li2022bevformer}. Recent approaches like StreamPETR \cite{wang2023exploring} utilize multi-view videos for training, emphasizing the need for extensive, well-annotated datasets. However, gathering and annotating such data across diverse conditions is challenging and costly. To address the mentioned challenges, recent advancements in generative models show that synthetic data can effectively improve performance in various tasks like object detection and semantic segmentation.

As the involvement of temporal data in video plays a crucial role in relative perception tasks, our focus in this paper shifts to generating high-quality realistic videos. Achieving real-world fidelity requires high visual quality, cross-view and temporal consistency, and precise controllability. Notice that the potential of recent methods are limited due to disadvantages in including low resolution, fixed aspect ratios, and inconsistencies in object shape and color. 
Inspired by the success of Sora’s performance in task of generating high-quality, temporally consistent videos, we adapt the Diffusion Transformer (DiT) for controllable multi-view video generation in our work.

Our proposed framework is among the first few works which propose to use DiT for video generation in driving scenarios, enabling precise content control by integrating bird's-eye view (BEV) layouts and scene text. Building on top of OpenSora \cite{opensora} architecture, our method embeds joint cross-attention modules to manage the scene text and instance layouts from bird's-eye views. Extending the ControlNet-Transformer \cite{chen2024pixart} approach for road sketches, we ensure multi-view consistency with parameter-free spatial view-inflated attention. For the aim of supporting multi-resolution generation, faster inference, and various video length, we utilize OpenSora’s training strategy and introduce a novel classifier-free guidance technique to enhance control and video quality.

\section{Methodology}
\label{sec:method}

\begin{figure*}[htbp]
\centering
\includegraphics[width=\linewidth]{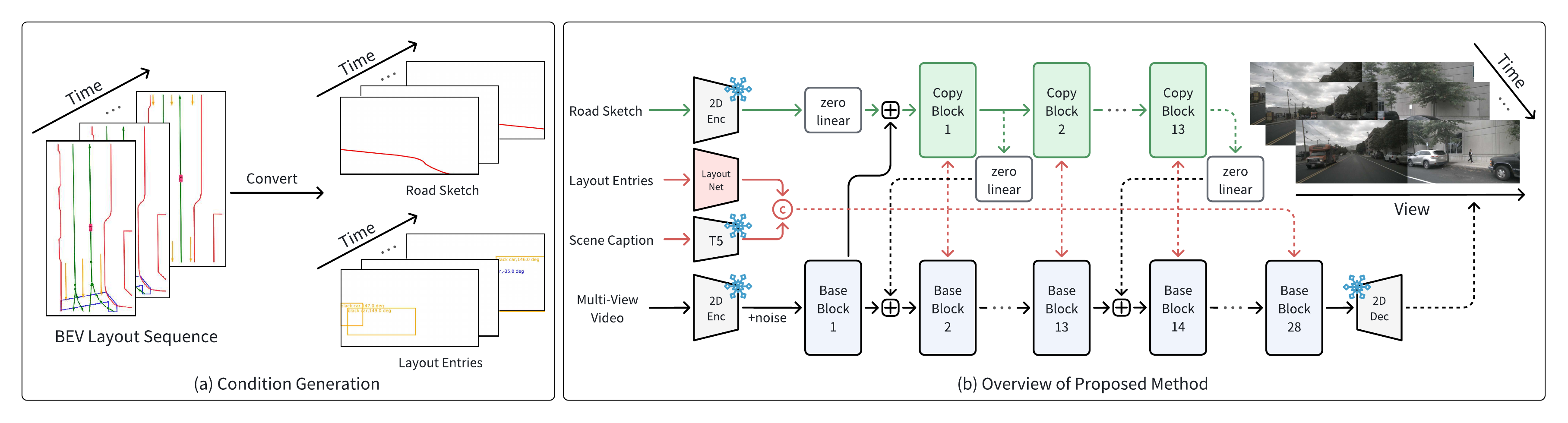}
\vspace{-2em}
\caption{The structural implementation with each individual components in our proposed method.}
\label{fig:first}
\end{figure*}


The overall architecture of our model is illustrated in Figure \ref{fig:first}. The parametric model proposed by OpenSora 1.1~\cite{opensora} is adopted as the baseline model. 
To achieve precise control over foreground and background information, we incorporate layout entries and road sketches, derived from 3D geometric data through projection, into the process of layout-conditioned video generation~\cite{ma2024unleashing}. Our proposed novel modules and training strategies will be introduced in the following sections accordingly.


\subsection{Mutli-Conditioned Spatial-Temporal DiT}
Following OpenSora 1.1~\cite{opensora}, we utilize a pre-trained and frozen variational autoencoder from LDM~\cite{rombach2022high} to extract latent features $z \in \mathbbm{R}^{V \times T \times 4 \times h \times w}$ from the input multi-view video clip, where $V$ represents the number of views, $T$ denotes the sequence length of frames, $h$ and $w$ denote the height and width of the latent features, respectively. These features are then modeled for spatiotemporal information using a 3D patch embedded. The textual input is encoded into 200 tokens using the T5 \cite{raffel2020exploring} language model.

\noindent \textbf{Spatial View-Inflated Attention.}
To guarantee the multi-view consistency during generation, we replace the commonly used cross-view attention modules~\cite{gao2023magicdrive,ma2024unleashing} with a parameter-free view-inflated attention mechanism. Specifically, we extend 2D spatial self-attention to enable cross-view interactions by reshaping the input from $B~\times~V~\times~T~\times H'~\times W'~\times~C$ to $B~\times~T~\times~(VH'W')~\times~C$ and treating $VH'W'$ as the sequence length. Consequently, our proposed approach improves the multi-view coherence without compensating with additional parameters.

\noindent \textbf{Caption-Layout Joint Cross-Atttention.}
Following MagicDrive~\cite{gao2023magicdrive}, we use a cross-attention mechanism to integrate scene captions and layout entries. Whereas the layout entries, \textit{i.e.} instance details such as 2D coordinates, heading and ID, are Fourier-encoded and combined into a unified embedding. Instance captions are encoded using a pre-trained CLIP~\cite{radford2021learning} model. These embeddings are concatenated and processed through an MLP, producing the final layout embedding, which, along with the scene caption embedding, conditions the cross-attention mechanism.

\noindent \textbf{ControlNet-Transformer.}
Delving into details, we introduce ControlNet-Transformer to ensure the precision towards the road sketch control inspired by PixArt-$\delta$~\cite{chen2024pixart}. 
Practically, a pre-trained VAE extracts the latent features from road sketches, which are then processed by a 3D patch embedder for the sake of consistency issue with our main network. To parameterize our mentioned design, 13 duplicated blocks are integrated with the first 13 base blocks with the DiT~\cite{peebles2023scalable} architecture. Each duplicated block combines the road sketch features and base block outputs, using spatial self-attention to reduce the computational overhead.
\begin{figure*}[t]
    \centering
    \includegraphics[width=0.99\linewidth]{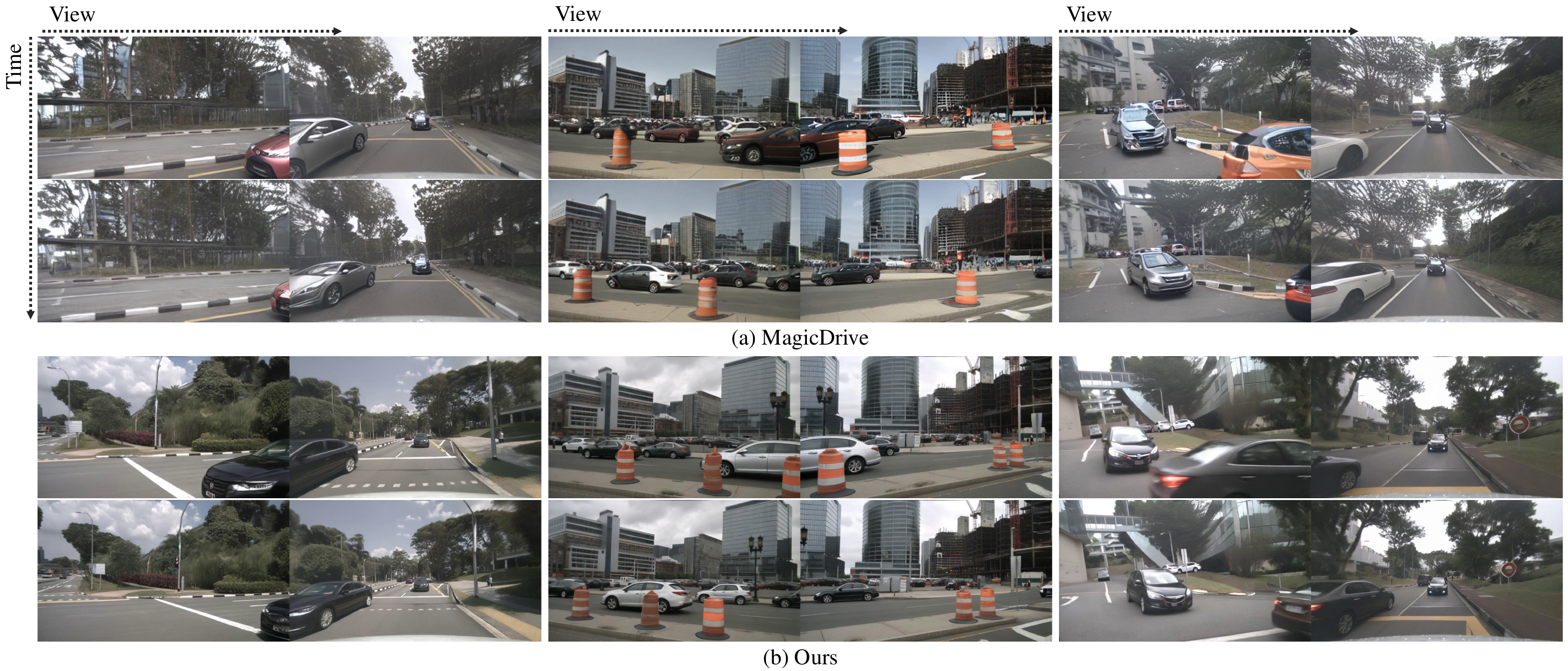}
    \vspace{-0.8em}
    \caption{Qualitative comparison of multi-view videos generated by our model and MagicDrive.}
    \label{fig:qualitative_cmp}
\end{figure*}

\begin{figure}[t]
    \centering
    \includegraphics[width=0.99\linewidth]{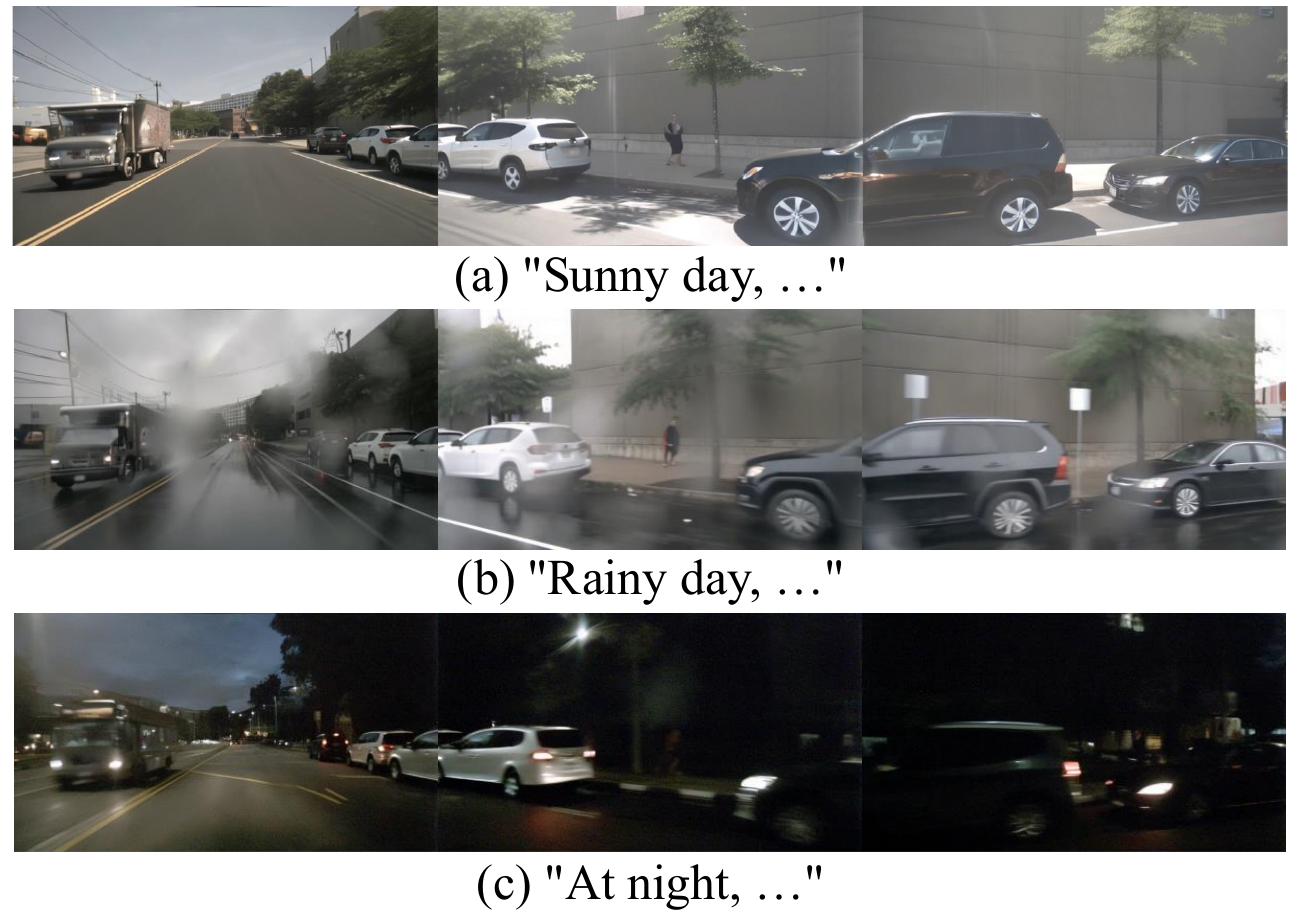}
    \vspace{-0.8em}
    \caption{The use-case of scene editing.}
    \label{fig:scene_edit}
\end{figure}

\subsection{Training}

\noindent \textbf{Variable Resolution and Frame Length.}
Following OpenSora~\cite{opensora}, we adopt the Bucket strategy, which ensures that videos within each batch have a consistent resolution and frame length.

\noindent \textbf{Rectified Flow.}
Inspired by OpenSora 1.2~\cite{opensora}, we replace IDDPM~\cite{nichol2021improved} with rectified flow~\cite{liuflow} during the later training stages for increased stability and reduced inference steps. Rectified flow, an ODE-based generative model, defines the forward process between data and normal distributions as
\begin{equation}
x_t=(1-t)x_0+tx_1 ~,
\end{equation}
\noindent where $x_1$ is a data sample, and $x_0$ is a sample from the normal distribution. The loss function is constructed as
\begin{equation}
\ell(\theta):= \mathbb{E}_{x_1,x_0}[\left \| \upsilon_{\theta}(x_t,t,c)-(x_1-x_0)  \right \|^2_2 ] ~,
\end{equation}
\noindent with $c$ encompassing the three conditions. 
Sampling is performed from $t=1$ to $t=0$ in $N$ steps via
\begin{equation}
x_{t-\frac{1}{N}} = x_t - \frac{1}{N} \upsilon_{\theta}(x_t, t, c), \forall t\in\left \{ 1,2,...,N \right \} /N ~.
\end{equation}

\noindent \textbf{First-$\bm{k}$ Frame Masking.}
To enable arbitrary-length video generation, we propose a first-$k$ frame masking strategy, allowing the model to seamlessly predict future frames from the preceding ones. Formally, given a binary mask $m$ indicating the frames to be masked—where the unmasked frames serve as the condition for future frame generation—we update $x_t$ as
\begin{equation}
x_{t} \leftarrow x_t \odot (1 - m) + x_1 \odot m ~,
\end{equation}
with losses calculated only on unmasked frames. During inference, video is generated autoregressively, with the last-$k$ frames of the previous clip conditioning the next.

\noindent \textbf{Classifier-free Guidance for Multi-Conditions.}
We observe that extending classifier-free guidance from the text condition to layout entries and road sketches enhances conditional control precision and visual quality. During training, we set the text condition $c_T$, layout condition $c_L$, and sketch condition $c_R$ to $\phi$ with a $5\%$ probability each, and also enforce a $5\%$ probability where all three conditions are simultaneously set to $\phi$. The guidance scales $\lambda_T$, $\lambda_L$, $\lambda_R$ correspond to the scene caption, layout entries, and road sketch, respectively, and measure the alignment between the sampling results and conditions. Inspired by~\cite{brooks2023instructpix2pix} , the modified velocity estimate is as follows:
\begin{align}
\upsilon'_\theta &= \upsilon_\theta(x_t, \phi, \phi, \phi) \\
&+\lambda_T\cdot(\upsilon_\theta(x_t, c_T, c_L, c_R)-\upsilon_\theta(x_t, \phi, c_L, c_R)) \\
&+\lambda_L\cdot(\upsilon_\theta(x_t, \phi, c_L, c_R)-\upsilon_\theta(x_t, \phi,\phi,c_R)) \\
&+\lambda_R\cdot(\upsilon_\theta(x_t, \phi, \phi, c_R)-\upsilon_\theta(x_t,\phi,\phi,\phi)) ~.
\end{align}





\section{Experiments}

\begin{table*}[htbp]
\vspace{-0.8em}
\centering
\begin{tabular}{ccccccc}
\hline
Method     & FVD$_\downarrow$   & Object mAP$_\uparrow$ & Map mIoU$_\uparrow$ & DTC$_\uparrow$ & CTC$_\uparrow$ & IQ$_\uparrow$ \\ \hline
MagicDrive & 221.90 & 11.73     & 18.44    & 0.8755    & 0.9251    & 48.85   \\
\textbf{Ours}       & \textbf{94.60} & \textbf{24.55}      & \textbf{35.96}    & \textbf{0.9132}    & \textbf{0.9446}    & \textbf{51.82}   \\ \hline
\end{tabular}
\caption{\label{tab:1}Quantitative comparison with MagicDrive. DTC, CTC and IQ represent DINO Temporal Consistency, CLIP Temporal Consistency and Imaging Quality, respectively. The best performances are presented in \textbf{bold}.}
\end{table*}

\subsection{Setups}

\noindent \textbf{Dataset and Evaluation Metrics.}
We train and evaluate our model using the nuScenes \cite{caesar2020nuscenes} dataset and the interpolated $12$Hz annotations provided by challenge. The generated multi-view videos are assessed based on distribution similarity (FVD), temporal consistency (DTC \cite{oquab2024dinov} and CTC \cite{radford2021learning}), visual quality (MUSIQ \cite{ke2021musiq}) and controllability. Controllability is evaluated through two perception tasks: 3D object detection and BEV segmentation, with BEVFormer \cite{li2022bevformer} serving as the perception model.

\noindent \textbf{Training Details.}
We train our method in four stages using eight NVIDIA A800 GPUs.
In the first stage, we fine-tune on OpenSora 1.1 \cite{opensora} checkpoints with fixed-resolution images of $512\times 512$ for $30k$ steps to control layout and sketch, training the ControlNet-Transformer, spatial attention, and layout net with spatial self-attention in base blocks.
In the second stage, we train the model for $26k$ steps with variable resolutions (144p, 240p, 360p) and frame lengths to adapt to the nuScenes dataset, continuing to use spatial self-attention.
The final two stages replace IDDPM with rectified flow, training for $20k$ steps at 144p to 360p, then $80k$ steps at higher resolutions (480p to full).

\noindent \textbf{Inference Details.}
We perform sampling inference using rectified flow with $30$ steps, choosing 480p resolution for a balance between inference time and visual quality. Each inference round uses a frame length of $16$. We set $\lambda_L$ and $\lambda_R$ to $2.0$, adjusting $\lambda_T$ to $1.0$ for night scenes and $7.0$ for other scenes to achieve the best results.

\subsection{Quality of Controllable Generation}

\noindent To assess the quality of generated videos, we compare our method with the challenge baseline, MagicDrive \cite{gao2023magicdrive}, using evaluations on 16-frame sequences. As shown in Table \ref{tab:1}, our model outperforms MagicDrive in terms of data distribution similarity, temporal consistency, visual quality, and controllability. Additionally, Figure \ref{fig:qualitative_cmp} illustrates that the videos produced by our model exhibit both higher visual quality and better spatial consistency. Figure \ref{fig:scene_edit} demonstrates the scene editing capability of our method, where the weather in the generated video changes according to the caption, while other objects remain unchanged.

\begin{table}[htb]
\vspace{-0.8em}
\scalebox{0.85}{
\begin{tabular}{ccccc}
\hline
Method          & FVD$_\downarrow$   & Object mAP$_\uparrow$ & Map mIoU$_\uparrow$ & Score$_\uparrow$  \\ \hline
CFG$_{T,L,R}$      & 94.60 & 24.55      & \textbf{35.96}    & \textbf{2.5962} \\
CFG$_{T,L}$        & 89.12 & 24.70      & 34.40    & 2.5487 \\
CFG$_{T}$          & \textbf{83.63} &  20.05          & 34.26         & 2.1749       \\
CFG$_{MagicDrive}$ & 164.48 & \textbf{26.18}     & 35.02    & 2.3618 \\ \hline
\end{tabular}
}
\caption{\label{tab:2}Ablation on the classifier-free guidance.}
\end{table}

\subsection{Ablation Study}

\noindent \textbf{Effect of Proposed Classifier-free Guidance.}
We compared different classifier-free guidance methods, both with and without unconditional layout and sketch considerations, as detailed in Table \ref{tab:2}. The "Score" is calculated as in the 1st round of the challenge, with CFG$_{T,L,R}$ being our proposed method. Excluding unconditional sketch (CFG$_{T,L}$) or both (CFG$_{T}$) yielded slightly better FVD but showed more pronounced differences in BEV segmentation and 3D object detection. We also evaluated CFG$_{MagicDrive}$ from MagicDrive \cite{gao2023magicdrive}, which performed well in controllability but had only satisfactory FVD. Ultimately, CFG$_{T,L,R}$ achieved the best overall score.

\section{Conclusion}
In this paper, we present the first DiT-based controllable multi-view video generation model tailored for driving scenarios. The integration of ControlNet-Transformer and joint cross-attention facilitates precise control over BEV layouts. Spatial view-inflated attention, combined with a comprehensive set of training and inference strategies, ensures high-quality and consistent video generation. Comparisons with MagicDrive and various visualizations further demonstrate the model’s superior control and consistency in generated videos.
{
    \small
    \bibliographystyle{ieeenat_fullname}
    \bibliography{main}
}


\end{document}